   \documentclass[final,10pt,twocolumn]{elsarticle}
 \usepackage[vmargin={20mm,20mm},hmargin={15mm,15mm}]{geometry}
 
\usepackage[utf8]{inputenc}
\usepackage{hyperref}

\usepackage{graphicx} 
\usepackage{amsmath}
\usepackage{amssymb} 
\usepackage{rotating}
\usepackage{caption} 
\usepackage{subcaption}

 \usepackage{lineno}

\DeclareMathOperator*{\argmin}{argmin}
\usepackage{ifthen}
 
 \usepackage[usenames,dvipsnames]{color}

 \journal{Pattern Recognition}

\begin{document}

\begin{frontmatter}

\title{Mapping and Localization from  Planar Markers}

\author{Rafael~Mu\~noz-Salinas\corref{cor1}\fnref{fn1}}
\ead{in1musar@uco.es}

\author{Manuel J. Mar\'in-Jimenez}
\ead{mjmarin@uco.es}

\author{Enrique Yeguas-Bolivar}
\ead{eyeguas@uco.es}

\author{R. Medina-Carnicer}
\ead{rmedina@uco.es}

\cortext[cor1]{Corresponding author}
\fntext[fn1]{Computing and Numerical Analysis Department, Edificio Einstein. Campus de Rabanales, C\'ordoba University, 14071, C\'ordoba, Spain, Tlfn:(+34)957212255}

\address{Computing and Numerical Analysis Department, C\'ordoba University, Spain}

\begin{abstract}
Squared planar markers are a popular tool for fast, accurate and robust camera localization, but its use is frequently limited to a single marker, or at most, to a small set of them for which their relative pose is known beforehand. Mapping  and localization from a large set of planar markers  is yet a scarcely treated problem in favour of keypoint-based approaches. However,  while keypoint detectors are not robust to rapid motion, large changes in viewpoint, or significant changes in appearance,   fiducial markers can be robustly detected under a wider range of conditions.  This paper proposes a novel method to simultaneously solve the problems of mapping and localization from a set of squared planar markers. First, a quiver of pairwise relative marker poses is created, from which an initial pose graph is obtained. The pose graph may contain small pairwise pose errors, that when propagated, leads to large errors. Thus, we distribute the rotational and translational error along the basis cycles of the graph so as to obtain a corrected pose graph. Finally,  we perform  a  global pose optimization  by minimizing the reprojection errors of the planar markers in all observed frames. The experiments conducted show  that our method performs better than Structure from Motion and visual SLAM techniques.
  \end{abstract}
\begin{keyword}
Fiducial Markers, Marker Mapping, SLAM, SfM.
\end{keyword}

\end{frontmatter}

 \section{Introduction}

Camera pose estimation is a common problem in several  applications such as robot navigation \cite{williams09,royer2007monocular} or augmented reality \cite{augmentedreality,artoolkit,lepetitpose}.  The goal of camera pose estimation   is to determine the three-dimensional position of a camera w.r.t. a known reference system.

\begin{figure}[t!]
	\centering
		\includegraphics[width=0.5\textwidth]{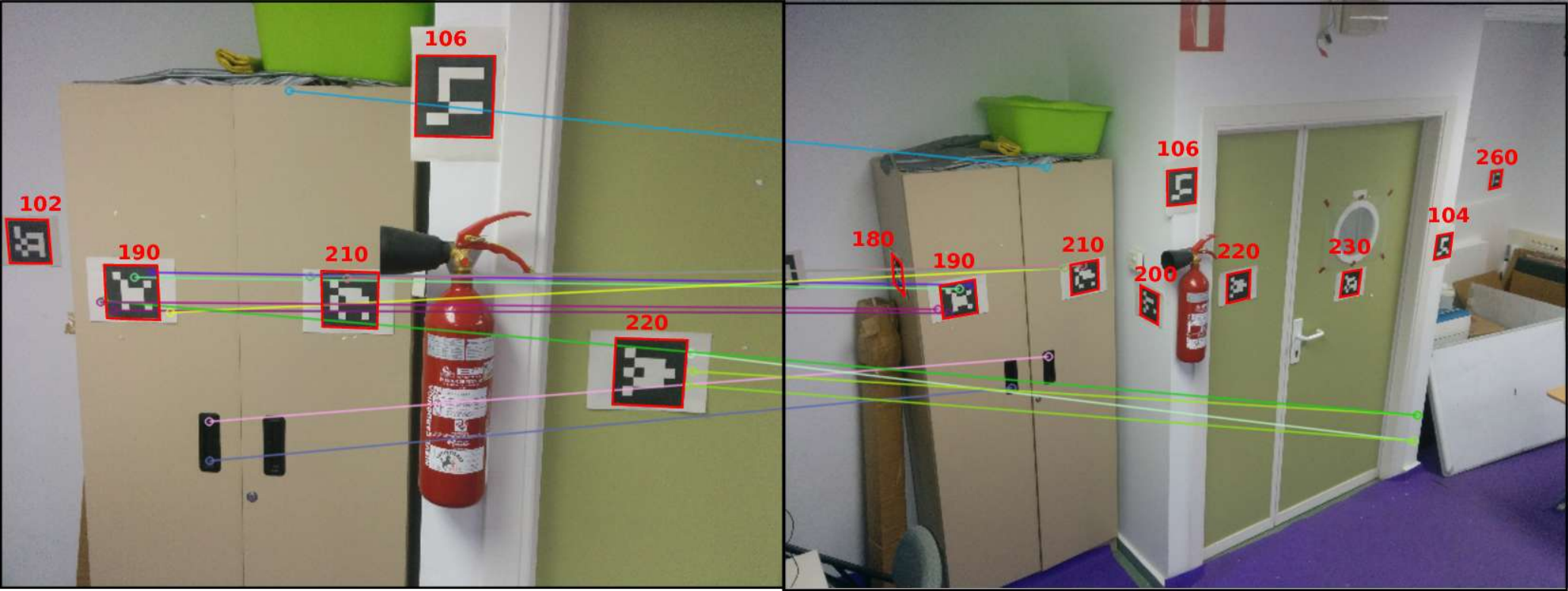}
	\caption{Example showing the matching capabilities of keypoints versus fiducial markers systems. Coloured lines show the best matches obtained by the SURF keypoint detector. Red rectangles show the markers detected along with its identification. Despite  large viewpoint changes, fiducial markers are correctly localized and identified. }
	\label{fig::surfvsaruco}
\end{figure}

To solve that problem, a great part of the research focuses on using natural landmarks, being  Structure from Motion (SfM) and  Simultaneous Localization and Mapping (SLAM), the two main approaches.  Both methods rely on keypoints \cite{bay2006surf,rosten2006machine,lowe2004distinctive}, which detect distinctive features of the environment. However, keypoint matching has a rather limited invariability to scale, rotation and scale, which in many cases makes them incapable of identifying a scene under different viewpoints. Thus, mapping an environment for tracking purposes under unconstrained movements requires a very exhaustive exploration. Otherwise, localization will fail from locations  different from  these employed for mapping. Take as example Fig.~\ref{fig::surfvsaruco}, where two images of the same scene are shown from different viewpoints and the SURF~\cite{bay2006surf} keypoint matcher is applied, showing as coloured lines the detected matches. Only two correct matches are obtained in this scene. 

Squared planar markers, however, are designed to be easily detected from a  wider range of locations \cite{Aruco2014,artagPAMI,artoolkit,studierstube,GarridoJurado2015}. Most frequently, squared markers use an external (easily detectable) black border and an inner binary code  for identification, error detection and correction. A single marker provides four correspondence points  which can be localized with subpixel precision to obtain an accurate camera pose estimation.  The scene in Figure~\ref{fig::surfvsaruco}  contains a set of planar markers which have been properly detected and identified  despite the viewpoint changes. However, camera localization from a planar marker suffers from the ambiguity problem~\cite{rpp:pami}, which makes it impossible to reliably distinguish the true camera location in some occasions.

Despite their advantages, large-scale mapping and localization from planar markers is a problem scarcely studied in the literature in favour of  keypoint-based approaches.  While it is true that some environments cannot be modified, in many occasions it is possible  to place as many markers as desired. In these cases, a large-scale and cost-effective localization system can be done using  planar markers exclusively. Additionally, in many indoor environments, such as labs or corridors, there are frequently  large untextured  regions from which keypoints can not be detected. If the environment must be texturized, then, it would be preferable to do it with fiducial markers, since they can be identified from a wider range of viewpoints than keypoints.

This work proposes a  solution to the problem of mapping and localization from planar markers. The  contribution  of this work is three-fold. First, we propose to tackle the marker mapping problem as a variant of the Sparse Bundle Adjustment problem, but considering that the four corners of a marker must be optimized jointly. As a consequence, our approach reduces the number of variables to be optimized and  ensures that the true distance between corners is enforced during optimization. Second, we propose a graph-based method to obtain the initial map of markers dealing with the ambiguity problem. To that end, we first create a quiver of poses from which an initial pose graph is obtained which is then  optimized  distributing the rotational and translational errors along its cycles.  Third, we propose a localization method considering all visible markers, which is able to cope with the ambiguity problem.

In order to validate our proposal, it has been evaluated against two SfM and two SLAM state-of-the-art methods, and the results  show that our proposal improves them.

The rest of this paper is structured as as follows. Section~\ref{sec::rel_works} explains the related works, while Sect.~\ref{sec::initial_concepts} presents some initial concepts and definitions. Later, Sect.~\ref{sec::proposed_solution} explains our proposal and Sect.~\ref{sec::experiments} the experiments conducted. Finally, Sect.~\ref{sec::conclusions} draws some conclusions.

\section{Related works}
\label{sec::rel_works} 
This section provides an overview of the main research related to ours.
\subsection{Structure from Motion}

Structure from Motion techniques take as input a collection of images of the scene to be reconstructed, from which keypoints are detected so as to create a connection graph. 
From the set of image matches,  the relative position of the cameras is obtained by either an incremental or a global approach.  Incremental approaches \cite{Bundler,vsfm}  select an initial good two-view reconstruction,  and images are repeatedly added along with their triangulated matched keypoints. At each iteration, bundle adjustment is applied to adjust both structure and motion. Global approaches \cite{Moulon:2013,wilson_eccv2014_1dsfm,Ozyesil_2015_CVPR}, however, create  a pose graph  by computing  pairwise view poses. In a first step, they compute the global rotation of the views, and in a second step the camera translations. All cycles of the graph imposes multi-view constrains that when enforced reduces the risk of drifting occurring in incremental methods. 
Both incremental and global approaches end with a bundle adjustment process to jointly optimize the motion and structure components.

In order to compute the relative pose between two views it is necessary to assume that the scene is locally planar \cite{6906584}, so that the homography can be computed \cite{faugeras}, or compute the essential matrix, which can model both planar and general scenes using the five-point algorithm~\cite{Nister}. However, in most cases, a relatively large number of matches between image pairs is required in order to obtain reliable solutions.

\subsection{Simultaneous Localization and Mapping}

SLAM is the process of localizing a robot while navigating  in the environment and building a map of it at the same time. While many different sensors can be integrated to solve that problem,  visual SLAM aims at solving the problem using visual information exclusively. In \cite{PTAM},  Klein and Murray  presented their PTAM system, in which two different threads running in parallel  create and update a map of keypoints. The work was pioneer since showed the possibility of splitting the tasks into two different threads achieving real-time performance. However, their keypoint descriptors did not consider the detection of large loops.
 
The recent work of Mur-Artal {\it et al.}~\cite{orb-slam} presents a keyframe-based SLAM method using ORB keypoints~\cite{orb}. Their approach operates in real-time and is able to detect  the loop closure and correct the poses accordingly.  Engel {\it et al.} \cite{engel14eccv} 
 proposed semi-dense  monocular visual SLAM solution called LSD-SLAM. In their approach,  scenes are reconstructed in a semi-dense fashion, by fusing spatio-temporally consistent
 edges. However, in order to solve the relocalization and loop-closure problems, they use keypoint features.

As previously indicated, systems based on keypoints pose several drawbacks.
Tracking loss typically fails with rapid motion, large changes in viewpoint, or significant appearance changes.

\subsection{Fiducial Squared Markers}
Fiducial marker systems are composed by a set of valid markers and an algorithm which performs its detection and identification.  In the simplest cases, points are used as  fiducial markers, such as LEDs, retroreflective spheres or planar dots \cite{leds1,leds2}. In these approaches, segmentation is achieved by using basic techniques over controlled conditions, but  identification  involves a more complex process.  Other works use planar circular markers where the identification is encoded in circular sectors or concentric rings \cite{circularmarkers1,circularmarkers2}, 2D-barcodes technology   \cite{cybercode,visualcode}  and even some authors have proposed markers designed using evolutionary algorithms  \cite{reactivision}.

However,  approaches based on  squared planar markers  are the most popular ones \cite{Aruco2014,artagPAMI,artoolkit,studierstube,GarridoJurado2015,artagvsartoolkitplus}. They are comprised by an external black border and an internal (most often binary) code to uniquely identify each marker.
 Their main advantage is that a single marker provides four correspondence points (its four corners), which are enough to do camera pose estimation. Detection of such markers is normally composed by two steps. The first one consists in looking for square borders, which produces a set of candidates that can be either  markers or background elements. In the second step,  each candidate is analysed to extract its binary code and decide whether it is a marker or part of the background.  
 
Selecting  appropriate binary marker codes for an application is of great relevance to reduce the chance of errors. Some authors have employed classic signal coding techniques \cite{matrix,artagPAMI,artoolkitplus}, others heuristic approaches \cite{Aruco2014,olson2011} and even Mixed Integer Linear Programming (MILP) \cite{GarridoJurado2015} has been used to obtain  optimal solutions.

\subsection{The ambiguity problem in Planar Pose Estimation}

In theory, the pose of a camera w.r.t. four non-linear and coplanar points can be uniquely determined. However, in practice, there is a rotation ambiguity that corresponds to an unknown reflection of the plane about the camera's $z$-axis  
\cite{Oberkampf1996495,rpp:pami,Collins2014}. This can happen not only  when imaging small planes, or planes at a distance significantly larger than the camera's focal length, but also for cases with wide angle lenses and close range targets. Most modern  algorithms \cite{rpp:pami,Collins2014} operate by providing the two possible solutions, and the reprojection error of each one of them. 

In most of the cases, the reprojection error of one solution is much lower than the reprojection error of the other one. Then, no ambiguity problem is observed and the correct solution is the one with lowest error. In other occasions, thought, both solutions have similar reprojection errors. In the absence of noise in the corner estimation, the solution with lowest error is always the correct one, but in realistic scenarios it can not be guaranteed. Thus, in practice, when the reprojection errors of the two solutions are very similar it is not safe to decide upon any of them.   Robust methods for mapping and localization using squared planar markers must take this problem into consideration.

\subsection{Mapping and Localization with Squared Planar Markers}

Large-scale mapping and localization from planar markers is a problem scarcely studied in the literature in favour of  keypoint-based approaches. The work of Hyon and Young~\cite{5333379} presents an approach to SLAM with planar markers. An  Extended Kalman-Filter (EKF) is used to  track a robot pose while navigating in a environment with some markers in it. As markers are found,  they are added to the map considering the  current robot pose along with the relative pose of the marker and the robot. Their approach, however, does not consider optimizing the estimated marker locations nor the ambiguity problem. A similar approach is presented in \cite{yamada2009study} for an autonomous blimp. 

The work of Klopschitz and Schmalstieg \cite{Klopschitz07automaticreconstruction} shows a system for estimating the 3d location of fiducial markers in large environments. In their work, a video sequence of the environment is recorded and camera position estimated using SfM  (with natural keypoints). Once  camera locations are accurately obtained, marker locations are obtained by triangulation. Our approach, on the other hand,
deals the dual problem of camera and marker localization jointly, without  the need of using SfM techniques. Their approach relies then on the correct functioning of a SfM method, that as we have already commented, is not always possible.

The work of Karam {\it et al.} \cite{shaya2012self} proposes the creation of a pose graph  where nodes represents markers and edges the relative pose  between them. The map is created in an online process, and edges updated dynamically. Whenever a pair of makers are seen in a frame, their relative position is updated and if it is better than the previous one, replaced. For localization, their approach selects, from the set of visible markers at that time, the one whose path to a origin node is minimum. Their approach poses several problems. First, they do not account for the ambiguity problem. Second, they do only consider for localization one marker from all visible ones. However, using  all visible markers at the same time can lead to better localization results. Third, their experimental results conducted does not really prove the validity of their proposal in complex scenes.

Finally, the work of Neunert {\it et al.} \cite{neunert2015open} presents a monocular visual-inertial EKF-SLAM system based on artificial landmarks. Again, a EKF is employed to do SLAM fusing information from the markers and an  inertial measurement  unit.

\section{Initial concepts and definitions}
\label{sec::initial_concepts}
This section explains some initial concepts and definitions that will be useful through the rest of the paper.

\subsection{Three-dimensional transforms and camera model}

Let us consider  a three-dimensional point $\mathbf{p_a=(x,y,z)}$ in an arbitrary reference system $a$. In order to express such point into  another reference system  $b$ it must undergo  a rotation followed by a translation. 
Let us denote  by
\begin{equation}
\zeta=(\mathbf{r},\mathbf{t})~| ~ \mathbf{r},\mathbf{t}\in \mathbb{R}^3,
\end{equation}
the  three rotational and translational components $\mathbf{r}$ and $\mathbf{t}$. Using Rodrigues' rotation formula, the  rotation matrix $\mathbf{R}$ can be obtained from $\mathbf{r}$ as:

\begin{equation}
\mathbf{R}=\mathbf{I}_{3\times3}+ \overline{\mathbf{r} }  \sin \theta + \overline{\mathbf{r} }^2(1-\cos \theta),
\end{equation}
where $\mathbf{I}_{3\times3}$ is the  identity matrix and $\overline{\mathbf{r}}$ denotes the antisymmetric matrix
\begin{equation}
\overline{\mathbf{r} }=\left[ \begin{tabular}{ccc}0&-$r_x$&$r_y$\\$r_z$&0&-$r_x$\\-$r_y$&$r_x$&0\\ \end{tabular}\right]
\end{equation}
Then, in combination with $\mathbf{t}$, the $4\times 4$ matrix 

\begin{equation}
 \gamma=\Gamma(\zeta)=\left[\begin{tabular}{cc} $\mathbf{R}$ &$\mathbf{t}^\top$\\ 0 & 1 \end{tabular}\right]
 \label{eq::gamma_f}
\end{equation}

can be used to transform the point from $a$ to $b$ as:

\begin{equation}
\label{eq::transform_3d}
\left[\begin{tabular}{c} $\mathbf{p}_b^\top$\\1 \end{tabular}\right]
=\gamma\left[\begin{tabular}{c} $\mathbf{p}^\top_a$\\1 \end{tabular}\right]
\end{equation}

To ease the notation,  we will  define the operator  ($\cdot$) to express:

\begin{equation}
\mathbf{p}_b= \gamma \cdot \mathbf{p}_a.
\end{equation}

A point $\mathbf{p}$ projects in the camera plane into a pixel $\mathbf{u}\in \mathbb{R}^2$. Assuming that the camera  parameters are know, the projection can be obtained as a function:
\begin{equation}
\label{eq::projection}
\mathbf{u}=\Psi(\delta,\gamma,\mathbf{p}),
\end{equation}
where $$\delta=(f_x,f_y,c_x,c_y,k_1,\ldots,k_n),$$  refers to the camera intrinsic parameters, comprised by the focal distances ($f_x,f_y$), optical center ($c_x,c_y$) and distortion parameters ($k_1,\ldots,k_n$). The parameter $\gamma$  represents camera pose from which  frame was acquired, i.e., the transform that moves  a point from an arbitrary reference system to the camera one.

\subsection{Clarification on the notation}

Along this paper,  the term $\gamma$ will be used referring to transforms moving points between different reference systems. To avoid confusions, we provide  a clarification on the most relevant terms employed.
\begin{itemize}
\item  $f^t$: frame. Image acquired by a camera at the time instant $t$.
\item   frs: frame reference system. Reference system centred in the camera origin when the frame was acquired.  Each frame has its own frame reference system.
\item mrs: marker reference system. Reference system centred in a marker. Each marker has its own mrs.
\item grs: global reference system. The common reference system w.r.t. which we desire to obtain all measures.
\item $\gamma_i$: mrs $\rightarrow$ grs. Transform points  from the reference system of marker $i$ to the global reference system.

\item $\gamma_{j,i}$: mrs $\rightarrow$ mrs. Transform points  from the reference system of marker $i$ to the reference system of marker $j$.

 \item $\gamma^t$: grs $\rightarrow$ frs. Transform points  from the global reference system to the reference system of frame $t$.

\item $\gamma_i^t$: mrs $\rightarrow$ frs. Transform points  from the reference system of marker $i$ to the reference system of frame $t$.

\item $\gamma_{j,i}^t$: mrs $\rightarrow$ mrs. Transform points  from the reference system of marker $i$ to the reference system of marker $j$, according to the observation of both in frame $t$.

\end{itemize}

In general, when using  transforms $\gamma$, the superscript    refers to frames, while the underscript  refers to markers.

\section{Proposed Solution}
\label{sec::proposed_solution}

This section explains the basis of our  approach for planar marker mapping and localization. As our first contribution, we formulate the problem as a  minimization of the reprojection error of the marker corners found in a set of frames (Sect.~\ref{sub_sec::problem_formulation}), obtaining a  non-linear equation that can be efficiently  minimized with the Levenberg-Marquardt  algorithm (LM) \cite{IMM2004-03215} using sparse matrices. While the problem resembles the Bundle Adjustment, our formulation reduces the number of variables by jointly optimizing the four corners of each markers. It also ensures that the real distance between the markers is enforced during the optimization. 

\begin{figure*}[t!]
	\centering
		\includegraphics[width=1\textwidth]{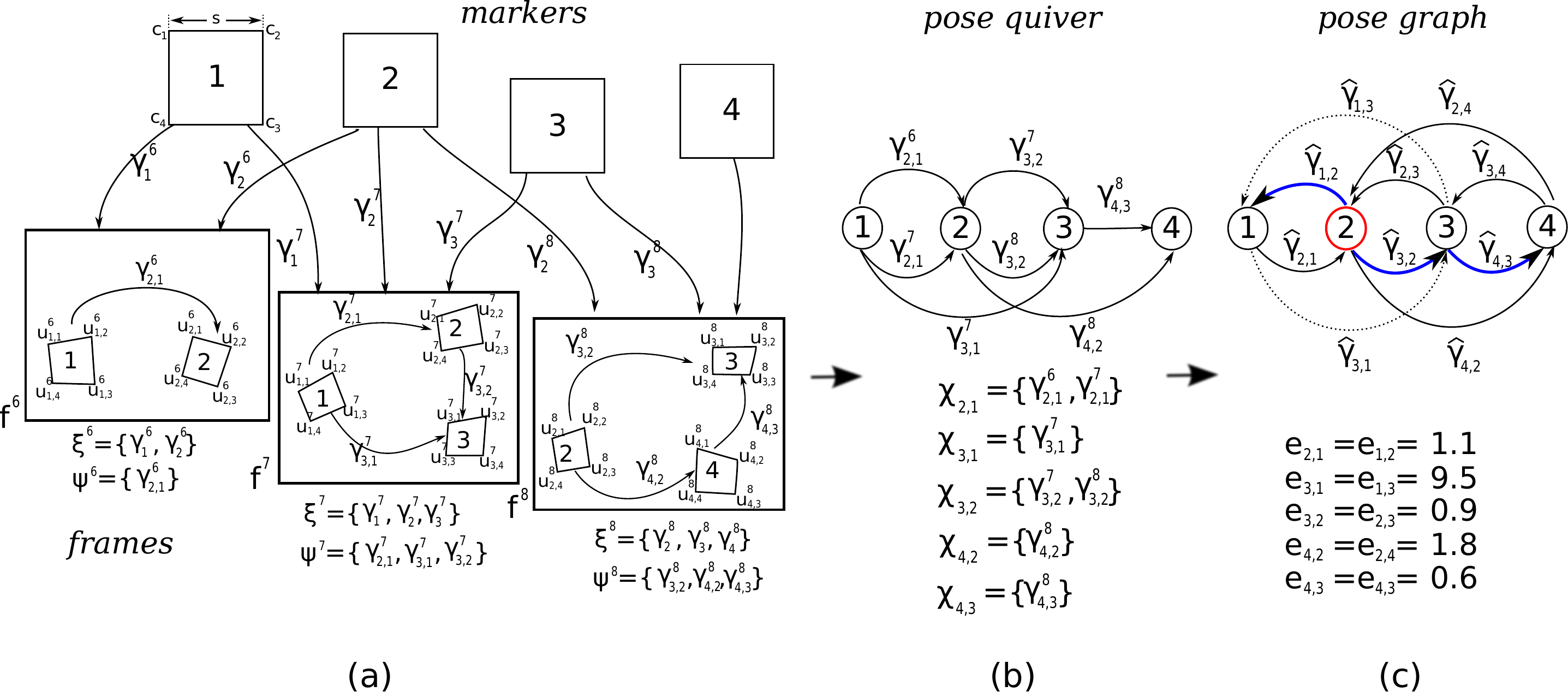}
	\caption{Figure summarizing the main concepts and variables of the proposed method. From the set of markers projected in the recorded frames, we obtain the pose quiver. Then, the best edges are obtained in order to create an initial pose graph that is further refined. Read text for further details.
	}
	\label{fig::overview_concepts}
\end{figure*}

Since the LM algorithm is a local search method, a good initial estimation  is required to avoid getting trapped in local minima. Obtaining an initial estimation for the marker poses is our second contribution.  Our idea is to create first a quiver where nodes represent markers, and edges their relative pose (Sect.~\ref{sec::pose_quiver}). When two markers are seen in a frame, their relative pose is computed, and an edge added to the quiver. The quiver is then employed to build an initial pose graph, where nodes represent markers, and edges between them represent the best relative pose observed (Sect.~\ref{sec::initial_pose_graph}).  This graph can be used to obtain an initial  approximation of the marker poses. However, it suffers from accumulative error as poses propagates along its nodes. We propose to distribute the errors along the graph cycles  \cite{pami:graph_error_mininization} obtaining a corrected version of the initial graph (Sect.~\ref{sec:pose_graph_optimization}). The corrected graph is then employed to obtain the initial estimation of the marker poses which serves as starting point for the LM optimization of Eq~\ref{eq::total_repr_err}. Nevertheless, it is yet required obtaining an initial approximation for the pose of the frames. In Sect.~\ref{sec::initial_frame_poses}, our third contribution is a  method to obtain the frame poses even in the presence  of erroneous solutions due to the ambiguity problem.  Figure~\ref{fig::overview_concepts} will help to clarify the concepts and notation through this Section.

\subsection{Problem formulation}
\label{sub_sec::problem_formulation}
Let us consider a squared planar marker, with sides  of length $s$, whose four corners can be expressed   w.r.t.   the marker center as: 
\begin{equation}
\begin{tabular}{l} 
$c_1$=( \begin{tabular}{rrr}~s/2,&-s/2,&0\end{tabular}),  \\
$c_2$=( \begin{tabular}{rrr}~s/2,&~s/2,&0\end{tabular}),\\
$c_3$=( \begin{tabular}{rrr}-s/2,&~s/2,&0\end{tabular}),\\
$c_4$=( \begin{tabular}{rrr}-s/2,&-s/2,&0\end{tabular}).\\ \end{tabular}
\end{equation}

We shall denote by 
\begin{equation}
\label{eq::marker_set}
    \mathcal{M}=\{m\},
\end{equation}
to the set of markers placed in the environment (each marker being uniquely identified), and   by $\gamma_{m}$ their poses, i.e., the transform that move points from the mrs to the grs.

Let us consider that a video sequence of the environment is recorded, and that a marker detector is applied to each  frame of the sequence.  Then, we shall denote by 
\begin{equation}
f^t=\{i~|~i\in\mathcal{M}\}
\end{equation}

to the set of markers detected in frame $t$ and by 

\begin{equation}
\label{eq::marker_2d_observations}
\omega_i^t=\{\mathbf{u}_{i,k}^t ~|~\mathbf{u}\in\mathbb{R}^2, k=1\ldots4\} \\
\end{equation}
the pixel locations in which the four corners of  marker $i$ are observed.  Please notice that for mapping purposes only frames observing at least two markers are considered, i.e.,  $|f^t|>1$. Also, we shall  use $\gamma^t$ referring to  the matrix that transforms a point from the  grs to the  frs of frame $f^t$.

In our problem, the poses $\gamma_m$ and $\gamma^t$ are parameters to be optimized, and $\omega_i^t$ the available observations. The camera parameters $\delta$ can also be included as part of the optimization process if desired. Then, the problem resembles the  Bundle Adjustment (BA) one. The main difference is that while BA assumes points to be independent from each other,  in our formulation, the four points of a marker forms a rigid object represented by only six parameters (i.e. $\gamma_m$). It brings two main advantages. First a reduction in the number of parameters, and thus, in the complexity of the optimization. Second, we  ensure that the distance between the consecutive corners is $s$, which would not be guaranteed using a general BA formulation.

In any case, the problem reduces to  minimize the squared reprojection error of the marker's corners in all frames so as to find the values of the parameters. The reprojection error of  marker $i$ detected  in frame $f^t$ is obtained comparing the observed projections of its corners $$\mathbf{u}_{i,j}^t, j=1...4, i\in f^t,$$  with the  predicted ones as:

\begin{equation}
e^t_i=\sum_{j} \left[  \Psi(\delta,\gamma^t, \gamma_i \cdot \mathbf{c}_j)-\mathbf{u}_{i,j}^t \right]^2,
\end{equation}
where $\Psi$ is the projection function defined in Eq.~\ref{eq::projection}.

Therefore, the total reprojection error in the whole set of frames is expressed as a function of the marker poses, frame poses, and   camera intrinsic parameters as:

\begin{equation}
\label{eq::total_repr_err_over}
\mathbf{e}(\gamma_1,\ldots,\gamma_M,\gamma^1,\ldots,\gamma^N,\delta)=\sum_t   \sum_{i\in f^t}  e^t_i,
\end{equation}
where $M$ and $N$ represent the number of markers and frames, respectively. 

Since the matrices $\gamma^t$ and $\gamma_i$ are overparametrized representations of the six degrees of freedom of a $SO(3)$ transform, we can use their equivalent representation  $\zeta$ to reduce the search space. Then the  optimization problem  of Eq.~\ref{eq::total_repr_err_over} is equivalent to:
\begin{equation}
\label{eq::total_repr_err}
\begin{split} 
f(x)=\mathbf{e}(\zeta_1,\ldots,\zeta_M,\zeta^1,\ldots,\zeta^N,\delta)=\sum_t   \sum_{i\in f^t}  \dot{e}^t_i,
\end{split}
\end{equation}
where
\begin{equation}
 \dot{e}^t_i=\sum_{j} \left[  \Psi(\delta,\Gamma(\zeta^t), \Gamma(\zeta_i) \cdot \mathbf{c}_j)-\mathbf{u}_{i,j}^t \right]^2,
\end{equation}
and $\Gamma$ is defined in  Eq.~\ref{eq::gamma_f}.

Then, our goal is to find the minimum of Eq.~\ref{eq::total_repr_err}. To that end, the LM algorithm \cite{IMM2004-03215}, a curve-fitting method that combines the gradient descent   and the Gauss-Newton methods to find the minimum of a non linear function $f(\mathbf{x})$, is employed. 

It is an iterative algorithm requiring an initial guess for the parameter vector $\mathbf{x}$, that at  each iteration, is replaced by a new estimate  $\mathbf{x} + \mathbf{p}$. 

Let the Jacobian of $f(\mathbf{x})$ be denoted  $J$, then the method searches in the direction given by the solution $\mathbf{p}$ of the equations
\begin{equation}
 (J^\top J+\lambda \mathbf{I})\mathbf{p}_k=-J^\top f(\mathbf{x}), 
\end{equation}
 
\noindent where $\lambda$ is a nonnegative scalar and $\mathbf{I}$ is the identity matrix. The  damping factor $\lambda$, is dynamically adjusted at each iteration. If the reduction of the error is large, a smaller value can be used, bringing the algorithm closer to the Gauss-Newton algorithm. However,  if an iteration gives insufficient reduction in the error, the parameter is increased, making the method more similar to the gradient descent.

Please notice that in our case, the Jacobian is sparse, since in general, only a small subset of the markers will project on each frame. Thus, we will take advantage of sparse matrices to speed up calculation.

Since the LM algorithm is a local search method, a good initial guess must be provided if the function to be minimized has more than one local minimum (as happen in our case). For the camera parameters, the calibration method proposed  in \cite{Zhang:2000:FNT:357014.357025} can be employed to obtain the initial estimates in an off-line process.  For the initial estimation of the marker and frame poses, we explain below our proposal.

%
\subsection{Pose quiver}
%

\label{sec::pose_quiver}
Let $Q$ be the quiver of poses where nodes represent markers and edges the relative pose between them.

We shall denote  

\begin{equation}
\label{eq::xidef}
 \xi^t=\{\gamma_i^t~ |~i\in f^t\},
\end{equation}

to the set of poses estimated for the markers detected in frame $f^t$ using a planar pose estimation method (such as \cite{rpp:pami} or \cite{Collins2014}). The element $\gamma_i^t\in\xi^t$:mrs $\rightarrow$ frs, represents  the transform that moves points from the mrs of marker $i$ to the   frs  $f^t$. 

As previously indicated,  planar pose estimators return two poses, one  correct, and another one that corresponds to the ambiguous solution. In most cases, the reprojection error of one solution is much larger than the reprojection error of the other, so there is not ambiguity problem, i.e., it is clear that the solution with the lowest reprojection error is the good one. However, in some  other cases, both reprojection errors are very similar. Then, it is difficult to know which solution is the good one. When it happens,  we  discard the  observation of that marker, i.e.,  no solution added to $\xi^t$ for that particular marker. Thus, $|\xi^t|\leq|f^t|$.

The elements in $\xi^t$ can be employed to obtain the relative poses between the markers observed. We shall define $\gamma_{j,i}^t$: (mrs$\rightarrow$mrs), as the pose that transform points from the reference system of marker $i$ to the reference system of marker $j$ according to the observation in $f^t$. It is calculated as:
\begin{equation}
 \gamma_{j,i}^t= \left(\gamma_{j}^t\right)^{-1}\gamma_{i}^t =\left[ \begin{tabular}{cc} $\mathbf{R}_{j,i}^t$ & $\mathbf{t}_{j,i}^t$\\0&1\\ \end{tabular} \right].
\end{equation} 

We shall then denote
\begin{equation}
\label{eq::interesting_poses}
\psi^t= \{\gamma_{2,1}^t,\gamma_{3,1}^t,\cdots,\gamma_{n,1}^t,\gamma_{3,2}^t,\cdots,\gamma_{n,n-1}^t\},
\end{equation}
with $n=|\xi^t|$,
to the set of all {\it interesting} combinations of such transforms. Since in our problem, the transform from $i$ to $j$ and its inverse can be easily calculated as: 

$$\gamma_{i,j}^t=\left( \gamma_{j,i}^t\right)^{-1},$$
for the sake of efficiency, we will only consider of interest the transforms $\gamma_{j,i}^t$ such that $i<j$.


Each element in $\psi^t$ will be an edge of the quiver $Q$. For several reasons (such as noise, camera movement, low  resolution, etc), the quality of the different edges will not the same.   Our goal  is to determine the best relative pose between each pair of markers observed, $\widehat{\gamma}_{j,i}$, so as to create an initial pose graph.   

The best edge between two nodes is the one that better explains the relative pose between the two markers in all frames where they are observed. Let us consider the example shown in  Fig.~\ref{fig::overview_concepts} and focus on markers $1$ and $2$ as observed in frames $f^6$ and $f^7$.  We define 
\begin{equation}
\mathbf{\varepsilon}_i^t=\sum_k \left[\Psi(\delta,\gamma_i^t,c_k)-\mathbf{u}^t_{i,k}\right]^2,
\end{equation}
as the reprojection error of marker $i$ in frame $f^t$ according to the  solution $\gamma_i^t$. The reprojection errors of the markers  $\mathbf{\varepsilon}_1^6,\mathbf{\varepsilon}_2^6,\mathbf{\varepsilon}_1^7$ and $\mathbf{\varepsilon}_2^7$  must be very close to zero (since they are the best solutions provided by the planar pose estimator).

Let us define
\begin{equation}
\label{eq::repj_transform}
\mathbf{\varepsilon}(\gamma_{j,i}^{t'},f^t)=\sum_k \left[\Psi(\delta,\gamma_j^t,\gamma_{j,i}^{t'}\cdot c_k)-\mathbf{u}^t_{i,k}\right]^2,
\end{equation}
as the reprojection error obtained by applying the relative pose $\gamma_{j,i}^{t'}$ to transform the points of marker $i$ to the reference system of marker $j$, and then projecting the transformed points using the pose $\gamma_j^t$ (being $t$ and $t'$ two different frames both seeing markers $i$ and $j$). For instance, in Fig.~\ref{fig::overview_concepts}, $\mathbf{\varepsilon}(\gamma_{2,1}^6,f^7)$ corresponds to transform the points of  marker $1$ to the reference system of marker $2$ using the relative pose of frame $f^6:\gamma_{2,1}^6$. Then, the transformed points are projected to the frame $f^7$ using $\gamma_2^7$ and the reprojection error computed. 

We consider that the pose $\gamma_{1,2}^6$ is better than $\gamma_{1,2}^7$ if  $\mathbf{\varepsilon}(\gamma_{2,1}^6,f^7)< \mathbf{\varepsilon}(\gamma_{2,1}^7,f^6)$. With this idea in mind, finding the best relative pose reduces to the problem of finding the one that minimizes the reprojection error in all other frames, i.e.,
\begin{equation}
\label{eq::optimal_intermarker_pose}
\widehat{\gamma}_{j,i}= \argmin_{ \gamma\in\chi_{j,i}}  \mathbf{e}_{j,i}(\gamma)
\end{equation}
\noindent where $\chi_{j,i}$ is the set quiver edges connecting nodes  $i$ and $j$, and 
\begin{equation}
\label{eq::repj_error_edge}
\mathbf{e}_{j,i}(\gamma) = \sum_{k\in\mathcal{F}_{i,j}} \mathbf{\varepsilon}(\gamma,f^k),
\end{equation}
is the sum of the reprojection errors in the set of frames $\mathcal{F}_{i,j}$ containing both markers $i$ and $j$.


\subsection{Initial pose graph}
\label{sec::initial_pose_graph}
Using the best intermarker poses from the quiver $Q$, we shall create the directed pose graph $G$, where nodes represent markers and edges their relative pose.
For each edge $e=(i,j)$, we shall define its weight 
\begin{equation}
 \varpi(e)=\mathbf{e}_{j,i}(\widehat{\gamma}_{j,i}),
\end{equation}
and its pose 
\begin{equation}
\psi(e)=\widehat{\gamma}_{j,i}.
\end{equation}
 
 While the quiver $Q$ only had the edges such that $i<j$, $G$ contains also edges such that $i>j$. These new edges are obtained considering that the reprojection error from node $i$ to $j$ is the same as the reprojection from $j$ to $i$:
\begin{equation}
 \mathbf{e_{i,j}}=\mathbf{e_{j,i}},
\end{equation}
 and that the inverse relative pose can be obtained as
\begin{equation}
 \widehat{\gamma}_{i,j}=\left( \widehat{\gamma}_{j,i}\right)^{-1}.
\end{equation}

Using $G$, an initial  estimation of the markers pose in a common reference system can be obtained as follows. First, select a starting node $a$ as the grs (i.e., $\gamma_a=\mathbf{I}_{4\times4}$), then compute the minimum spanning tree ({\it mst}) of the graph. Given the path  $(a,b,\cdots,h,i)$ of the {\it mst} from node $a$ to node $i$, its pose in the grs  can be obtained as:

\begin{equation}
\label{eq::pose_composition_ingraph}
  \widehat{\gamma}_i=   \widehat{\gamma}_{a,b} \ldots    \widehat{\gamma}_{k,h} \widehat{\gamma}_{h,i}.
\end{equation}

The choice of the starting node is important since it conditions the quality of the poses, thus, it must be chosen appropriately.  We define the cost of a {\it mst} as the sum of the reprojection errors of all its edges. Then, we  select as starting node the one that minimizes such cost. This operation can be efficiently computed using the Floyd's algorithm~\cite{Floyd:1962:A9S:367766.368168}. In the example of Fig~\ref{fig::overview_concepts}, we show in red the starting node, and with blue lines the edges of the best  {\it mst}. In this particular example, the pose of the different  markers can be obtained as:
\begin{equation}
\begin{tabular}{l}
$ \widehat{\gamma}_1=\widehat{\gamma}_{1,2}$\\
$ \widehat{\gamma}_2=\mathbf{I}_{4\times4}$\\
$ \widehat{\gamma}_3=\widehat{\gamma}_{3,2}$\\
$ \widehat{\gamma}_4=\widehat{\gamma}_{4,3}\widehat{\gamma}_{3,2}$\\
\end{tabular}
\end{equation}

In any case, the initial poses estimated from the best  {\it mst} are not yet good initial approximations for the  optimization problem of Eq.~\ref{eq::total_repr_err}, since the errors between markers incrementally propagate along the path from the starting node. Therefore, we will optimize the graph as explained below.

\subsection{Pose graph optimization}
\label{sec:pose_graph_optimization}
The graph  $G$ may contain errors in the relative poses that when propagated along a path can lead to large final errors, specially for the markers that correspond to the leaf nodes of the tree.  Our goal here is to obtain a graph $\widetilde{G}$ where the relative poses of $G$ have been improved. To do so, we will propagate errors along the cycles of the graph~\cite{pami:graph_error_mininization}. 

In a first stage, we  remove outliers connections from the graph $G$ to prevent them from corrupting the  optimization. To do so, we compute the mean  and standard deviation of the weights for the edges   in the  {\it mst}. For the rest of the edges (not in the {\it mst}), we remove from the graph those  outside a $99$\% confidence interval on the mean (i.e., $2.58$ times the standard deviation around the mean). As a result, we obtain the subgraph $G'$ that will be employed for further optimization. In the example  of Fig.~\ref{fig::overview_concepts}(c), the connections between nodes $1$ and $3$ are drawn with dashed lines  indicating that they are outliers and thus not included in $G'$. Then, we propagate the errors along the cycles of $G'$ so as to obtain $\widetilde{G}$.

Let us consider a cycle $c=(1,\ldots,n)$ of graph nodes, i.e., a path of nodes starting and ending in the same node.  The optimal intermarker poses $\widetilde{\gamma}$ of the cycle should meet the following conditions:
\begin{itemize} 
 \item  The cycle must be consistent, i.e, the composition of transform matrices along the cycle must be the identity:
\begin{equation}
\mathbf{I}_{4\times4}=\widetilde{\gamma}_{1,2}\widetilde{\gamma}_{2,3}\ldots\widetilde{\gamma}_{n-1,n} \widetilde{\gamma}_{n,1} 
\end{equation}

 \item Assuming that the initial relative poses are relatively correct,  we should minimize the weighted squared error of the new and old poses:

\begin{equation}
 min \sum_{k\in c} w_{k,k+1}||\widetilde{\gamma}_{k,k+1}-\widehat{\gamma}_{k,k+1}||,
 \end{equation}
where the weight  $w_{k,k+1}$ is the confidence of each relative pose. It is defined as a value in the range $[0,1]$ inversely proportional to the reprojection error of the edge as:
\begin{equation}
\label{eq::node_weight}
w_{k,k+1}=\frac{1}{\mathbf{e}_{k,k+1} \sum_{k\in c} 1/\mathbf{e}_{k,k+1}}, 
\end{equation}
\noindent Equation~\ref{eq::node_weight}  is such that the weights of the edges in the cycle sum up to one, i.e., $$ \sum_{k\in c} w_{k,k+1}=1.$$
\end{itemize}

For the sake of simplicity, the rotational and translational components will be optimized separately \cite{pami:graph_error_mininization}. To do so, it will be necessary to decouple rotation from translation as will be explained later. But let us first  focus on the rotational components for a single cycle and denote by $\mathbf{E}_{k,k+1}$  the error rotation matrix  such that
 
\begin{equation}
 \mathbf{R}_{1,2}\cdots\mathbf{R}_{k,k+1}\mathbf{E}_{k,k+1}\mathbf{R}_{k+1,k+2}\cdots\mathbf{R}_{n,1}=\mathbf{I}_{3\times3}.
\end{equation}
This rotation matrix corrects the accumulated error of the cycle when moving between nodes $k$ and $k+1$. It can be proved that  matrix  $\mathbf{E}_{k,k+1}$ can be broken up into fractional portions of the whole rotation thus defining 
\begin{equation}
\mathbf{E}_{k,k+1}^{\alpha_{k,k+1}}=\exp\{\alpha_{k,k+1} \ln \mathbf{E}_{k,k+1}\},
\end{equation}
which shares the same axis of rotation as $\mathbf{E}_{k,k+1}$ but the angle of rotation has been scaled by $\alpha_{k,k+1}$. Then, it is possible to distribute the error along the elements of the cycle by computing the corresponding error matrix between each one of the nodes. As a consequence, the optimal rotation matrices can be obtained as:
\begin{equation}
 \widetilde{\mathbf{R}}_{k,k+1}=\mathbf{E}_{k-1,k}^{\alpha_{k-1,k}} \mathbf{R}_{k,k+1}=\mathbf{R}_{k,k+1}\mathbf{E}_{k+1,k}^{\alpha_{k,k+1}},
\end{equation}
and the corrected cycle as:
\begin{equation}
 \mathbf{R}_{1,2}\mathbf{E}_{1,2}^{\alpha_{1,2}}\cdots\mathbf{R}_{k,k+1}\mathbf{E}_{k,k+1}^{\alpha_{k,k+1}}\cdots\mathbf{R}_{n,1}\mathbf{E}_{n,1}^{\alpha_{n,1}}=\mathbf{I}_{3\times3}.
\end{equation}

The parameter $\alpha_{k,k+1}\in[0,1]$ indicates the influence of an edge in the error distribution. High values indicate that the confidence in the edge is low, so that it requires a higher degree of correction, and vice versa. The value is computed as 
\begin{equation}
 \alpha_{k,k+1}=\frac{ 1/w_{k,k+1}}{\sum_{j\in c} 1/w_{j,j+1}}
\end{equation}
so that they sum up to one: $\sum_{k\in c}\alpha_{k,k+1}=1$.

In our case, substituting from Eq.~\ref{eq::node_weight}, we obtain:
\begin{equation}
 \alpha_{k,k+1}=\frac{ \mathbf{e}_{k,k+1} }{\sum_{j\in c}  \mathbf{e} _{j,j+1} },
\end{equation}
\noindent indicating that the smaller the reprojection error of an edge, the smaller the correction it requires. 
\begin{figure}[t!]
	\centering
		\includegraphics[width=0.15\textwidth]{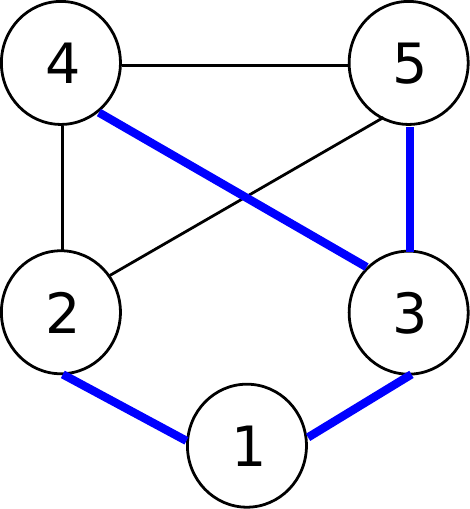}
	\caption{Basis cycles example. Let the minimum expanding tree be comprised by the edges set in blue lines, while the black lines are unused edges. Then, the  basis cycles of the graph are $\{ (1,2,4,3),(1,2,5,3),(3,4,5)\}$. }
	\label{fig::basic_cycles}
\end{figure}

Previous explanation has shown how to optimize the rotations of  a single cycle. For the whole graph, we will employ its {\it basis cycles}. Given the  {\it mst} of a graph, which contains all its nodes, adding a single unused edge  generates one basis cycle. The set of all basis cycles generated this way forms a complete set of basis cycles.  Figure~\ref{fig::basic_cycles}, showing a graph of five nodes, aims at clarifying this idea. Blue lines  represent the edges included in the {\it mst}, while the black ones are not. Then, the basis cycles are obtained by adding unused edges, e.g., the addition of edge $(2,4)$ creates the cycle $(1,2,4,3)$.  In total, this graph has the following three basis cycles: $\{ (1,2,4,3),(1,2,5,3),(3,4,5)\}$.

It is known from graph theory that  any circuit in a graph can be obtained as a linear combination of basis cycles in the edge space of the graph~\cite{pami:graph_error_mininization}. The distribution of the rotational error along  the cycles of the graph is achieved  by distributing the errors independently in each cycle, and then averaging the rotation estimates for edges appearing in more than one cycle. This process is repeated until convergence. 

Once the optimal rotations have been obtained, it is required to decouple translations from rotations before their optimization.  The decoupled translations are obtained by selecting a decoupling point, which will serve as the center of rotation between both markers.  In our case, the decoupling point  selected is the center between the two markers. The decoupled translation is then obtained as:
\begin{equation}
 \widehat{\mathbf{t}}^\prime_{1,2}=\left(\widehat{\mathbf{R}}_{1,2}-\widetilde{\mathbf{R}}_{1,2}\right)\mathbf{c}_2+\widehat{\mathbf{t}}_{1,2},
\end{equation}
where $\mathbf{c}_2$ is the decoupling point expressed in the reference system of node $2$.

Minimizing the translational error consists in finding the optimal estimates $\widetilde{\mathbf{t}}_{k+1,k}$ that are as close as possible to the  decoupled  values $\widehat{\mathbf{t}}^\prime_{k+1,k}$, but satisfy the constrain that any point $\mathbf{p}_k$ will map back to $\mathbf{p}_k$ as we compose the transformations about the cycle.
It is then the problem of finding the
 $$\min \sum ||\widetilde{\mathbf{t}}-\widehat{\mathbf{t}}^\prime||$$
 subject to
\begin{equation}
\label{eq::trans_error_cycle}
\begin{split} 
 \mathbf{p}_k=\widetilde{\mathbf{R}}_{k,k+1}(\widetilde{\mathbf{R}}_{k+1,k+2}(\cdots(\widetilde{\mathbf{R}}_{k-1,k}\mathbf{p}_k+\widehat{\mathbf{t}}^\prime_{k-1,k})\cdots )\\
 +\widehat{\mathbf{t}}^\prime_{k+1,k+2}+\widehat{\mathbf{t}}^\prime_{k,k+1}
\end{split}
\end{equation}

In order to distribute the translation error along all the cycles of the graph, we obtain a set of constrain equations (one per cycle)
that can be derived from Eq.~\ref{eq::trans_error_cycle} as:

\begin{equation}
 \begin{tabular}{ccccc}
  $M_{a_1,a_2}\widehat{\mathbf{t}}^\prime_{a_1,a_2}$&+$\cdots$+&$M_{a_k,a_1}\widehat{\mathbf{t}}^\prime_{a_k,a_1}$&$=$&$0$ \\
  $M_{b_1,b_2}\widehat{\mathbf{t}}^\prime_{b_1,b_2}$&+$\cdots$+&$M_{b_k,b_1}\widehat{\mathbf{t}}^\prime_{b_k,b_1}$&$=$&$0$ \\
     & $\cdots$ &   & \\
  $M_{n_1,n_2}\widehat{\mathbf{t}}^\prime_{n_1,n_2}$&+$\cdots$+&$M_{n_k,n_1}\widehat{\mathbf{t}}^\prime_{n_k,n_1}$&$=$&$0$ \\
 \end{tabular}
\end{equation}

This is a quadratic minimization problem with linear constrains that can be solved using Lagrange multipliers.

The correction of the pairwise translational and rotational errors of the graph $G'$ leads to the optimized graph $\widetilde{G}$ from which an initial marker poses can be obtained   w.r.t.  the  staring  node of the  {\it mst} as indicated in Eq.~\ref{eq::pose_composition_ingraph}. These poses are then further refined by jointly minimizing the reprojection error of all markers in all frames by   minimizing Eq.~\ref{eq::total_repr_err}.  

However, in order to fully solve the optimization problem, it is also required to obtain an initial estimation of the frame poses $\gamma^t$ as will be explained below.

\subsection{Initial frame pose estimation}
\label{sec::initial_frame_poses}
The initial  pose of the frames must be estimated considering that the individual marker pose estimation is subject to the ambiguity problem. We propose here a method to provide an initial estimation that is correct even if  ambiguity occurs.  Let us denote by:

\begin{equation}
 \Theta^t=\{\gamma^t_i,\dot{\gamma}^t_i\}~\forall~i\in f^t,
\end{equation}
to the set of poses computed by a planar pose estimator method for the  markers observed in frame $f^t$. For each marker $i$, the estimator provides two solutions, ${\gamma}^t_i$ and $\dot{\gamma}^t_i$, thus,   $|\Theta^t|=2|f^t|$. The reprojection error of the first solution is lower than than the reprojection error of the second. However, if the difference is  small, then, we have the ambiguity problem, i.e., it is not possible to decide which is the correct solution.

An estimation of the pose frame $\gamma^t$ can be calculated from each element in $\Theta^t$ as:
\begin{equation}
 \widehat{\gamma}^t=\gamma^t_i  (\widetilde{\gamma}_i)^{-1},
\end{equation}
where $\widetilde{\gamma}_i$ is obtained from  $\widetilde{G}$. Then the problem  becomes the one of finding the best estimation $\widetilde{\gamma}^t$ from the set of elements in $\Theta^t$. 

As for the quiver, the best estimation is the one that minimizes the reprojection error for all the markers observed in $f^t$. Thus, we define:
\begin{equation}
 \mathbf{e}(\gamma^t_i)= \sum_{j\in f^t} \sum_k \left[ \Psi\left(\delta,\gamma^t_i(\widetilde{\gamma}_i)^{-1}, c_k \right) -\mathbf{u}^t_{j,k}\right]^2,
\end{equation}
as the sum of reprojection errors of all markers observed in $f^t$ when the estimation $\gamma^t_i\in\Theta^t$ is employed. In this case, if 
$\gamma_i^t$ is an erroneous pose (because of the ambiguity problem), it only obtains low reprojection error for marker $i$, but not for the rest of markers. However, a good solution  obtains low reprojection errors for all the markers in $f^t$. Thus, the  best initial position reduces to find:
\begin{equation}
\label{eq::frame_pose_estimation}
\widetilde{\gamma}^t= \argmin_{\gamma^t_i\in\Theta^t} \mathbf{e}(\gamma^t_i).
\end{equation}

The poses obtained using Eq.~\ref{eq::frame_pose_estimation} can be used as initial solutions for the optimization of Eq.~\ref{eq::total_repr_err}.

\section{Experiments}
\label{sec::experiments}

This section explains the experiments carried out to validate our proposal using seven different experiments.  The source code and video sequences employed for our experiments are  publicly \href{http://www.uco.es/grupos/ava/node/25}{available} \footnote{\url{http://www.uco.es/grupos/ava/node/25}}.   
All the tests  were run on a $i7$ Intel computer with $8$Gb of RAM, and our code is parallelized in the parts where it can be. The ArUco library  \cite{Aruco2014,GarridoJurado2015} was employed for marker detection in the video sequences recorded and also for calibrating the cameras employed.

The first five experiments tests   compare the results of our method with those provided by the VisualSFM \cite{vsfm} and OpenMVG \cite{Moulon:2013} tools. The first one implements an incremental approach, while the second one implements a global approach. Both pieces of software automatically find keypoints between the images and then are able to find both the 3D location of the points and the pose of the cameras. For this work, we are only interested in the second phase of the process. Thus,  we provided the matches and camera intrinsic parameters to the tools so that only the SfM algorithm is employed. An advantage of this way of working is that there are not incorrect matches and the results can be compared with our method. The sixth experiment compares our approach with two state-of-the-art SLAM methods: namely LSD-SLAM\cite{engel14eccv} and ORB-SLAM \cite{orb-slam}. In all cases, we employed the implementation provided by the authors.  The final test presents a reconstruction example using a minimal set of images, showing the method's capability to be employed as a cost-effective localization system with very few images of the  environment.

Two different measures can be obtained to evaluate the quality of the proposed approach: the accuracy in the estimation of the marker poses,  and the accuracy in the estimation of  the frame poses. The first one can be evaluated by calculating  Absolute Corner Error (ACE), computed as the root mean squared error (RMSE)  between the estimated three-dimensional marker corners locations and the ground truth ones.  In order to to this, it is necessary  to transform the estimated corners to the  ground truth reference system, which can be done using  Horn's method~\cite{Horn:87}.  The accuracy of the  estimated frame poses is obtained using the Absolute Trajectory Error (ATE) measure, which calculates the RMSE between the translation components of the frame poses estimated and the ground truth ones.

\begin{figure}[t!]
	\centering
		\includegraphics[width=0.5\textwidth]{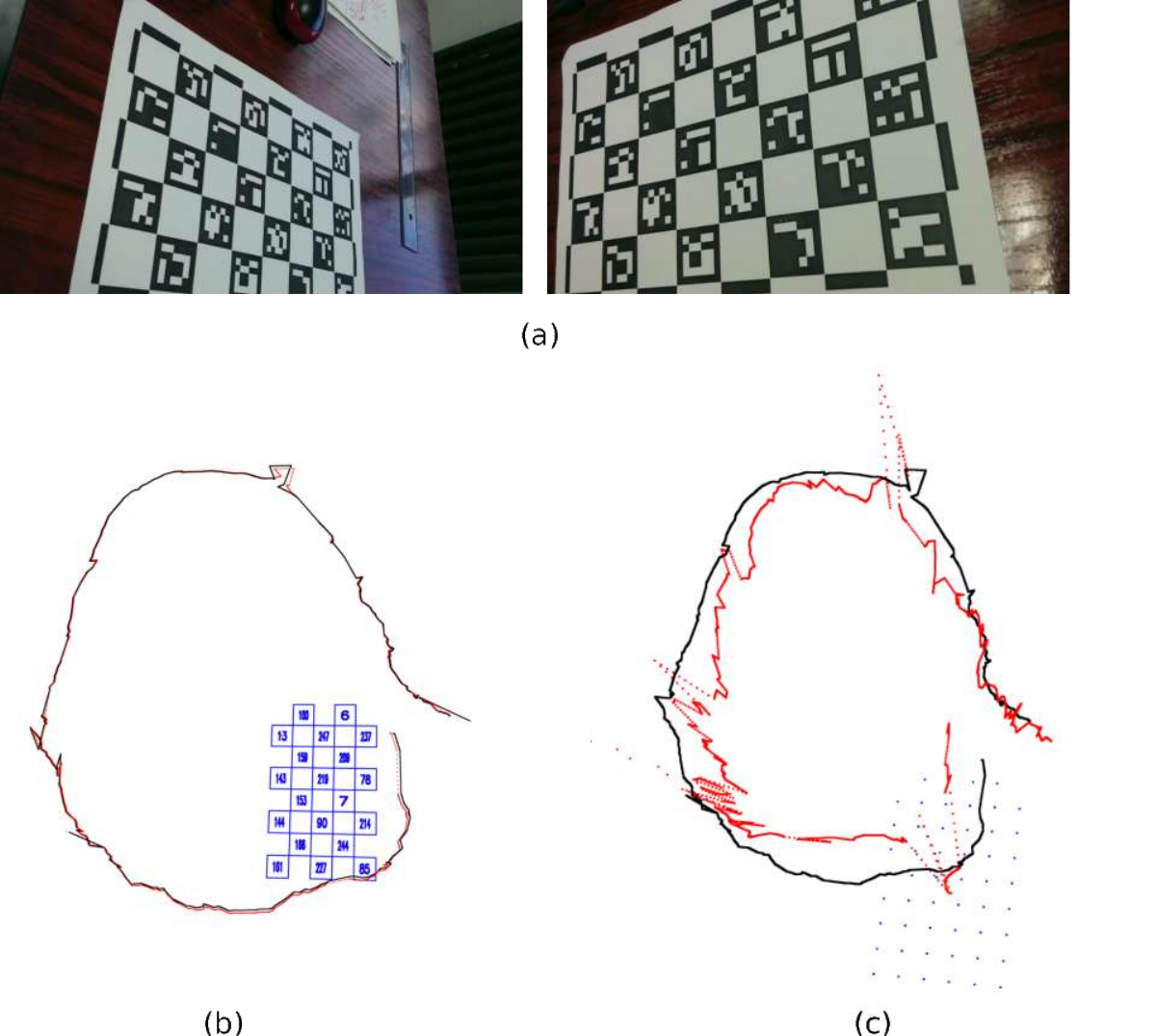}
	\caption{Results obtained for the calibration pattern. (a)  Snapshots of the image sequence. (b) Three-dimensional reconstruction obtained with our method. Markers (in blue), estimated trajectory in red, and ground truth trajectory in black. (c)  Three-dimensional reconstruction obtained with VisualSFM along with the estimated and ground truth trajectories.}
	\label{fig::pattern_test}
\end{figure}
\subsection{First test: Calibration board}

In our first test, we employ the calibration board  provided by the ArUco library for marker detection and calibration \cite{Aruco2014,GarridoJurado2015}. It was printed on a $A4$ piece of paper, with $20$ markers  of size $3.25$~cm (see Fig.~\ref{fig::pattern_test}(a)).

A video sequence of $523$ frames  was recorded using the camera in the Nexus-5 mobile phone  (with a resolution of $1920\times1080$  pixels) at a distance of approximately $50$~cm around the board.   Both the location of  the markers corners and the frame poses are obtained  using the ArUco library and used as ground truth. The computing time, ACE and ATE measures obtained by  the different methods for this sequence are shown in Table~\ref{tab::test_one_table}.
\begin{table}
\begin{center}
 \caption{ Results of the different methods for the calibration pattern sequence of the first test. For all measures, the lower the values, the better the results of  the method. }
\begin{tabular}{ l|r|c|c}
 \label{tab::test_one_table}

  \textit{Method}   & Comp. Time & ACE  & ATE \\
  \hline
  Ours & $14$~secs & $0.48$~mm & $4.32$~mm \\
  VisualSFM & $123$~secs & $0.64$~mm &$0.11$~m \\
  OpenMVG & $ 1211$~secs & $0.45$~mm& $2.24$~mm \\
\end{tabular}
\end{center}
\end{table}
Our method required $14$~secs to  process the sequence, excluding the time required to detect and identify the markers. In this test, the initial pose graph is a completely connected one, since there are frames in which all markers are seen simultaneously. As a consequence, the initial pose graph obtains very good initial estimates and the subsequent optimization does not produce significant improvements.  The average reprojection error obtained after the  final optimization is $0.7$ pixels.  Figure~\ref{fig::pattern_test}(b) shows the three-dimensional reconstruction obtained along with the trajectory of the camera. The red line represents the trajectory computed with our method (frame poses), and the black one is the ground truth.

\begin{figure}[t!]
	\centering
		\includegraphics[width=0.5\textwidth]{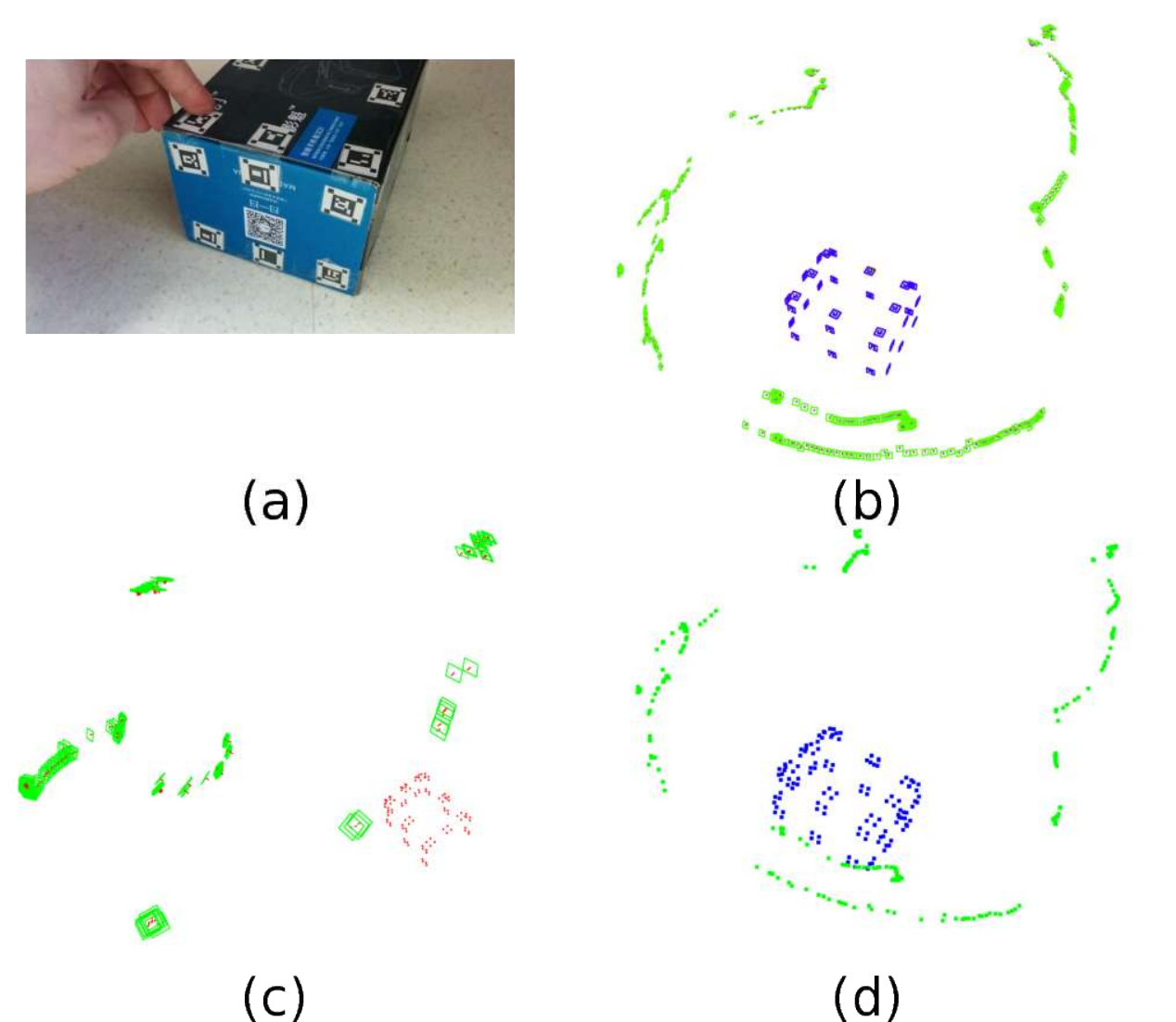}
	\caption{Results obtained for the cardboard box. (a)  Snapshopt of the image sequence. (b) Three-dimensional reconstruction obtained with our method. (c)  Three-dimensional reconstruction obtained with VisualSFM. (d) Three-dimentional reconstruction with OpenMVG}
	\label{fig::arcube_test}
\end{figure}
Fig.~\ref{fig::pattern_test}(c) shows the reconstruction results obtained with VisualSFM, which took $123$~secs of computing time. Since the SfM returns the results up to a scale factor, it is necessary first to scale the results in order to compare them with the ground truth, and then using Horn transform to set a common reference system.  As can be observed, while the reconstruction of the points seems correct ($0.64$~millimeters), the estimation of the frame poses are not so precise. The ATE obtained by this method was $0.11$~meters.

Finally, the OpenMVG method obtained the most precise ATE and ACE, but at the expenses of a computing time two orders of magnitude higher than ours.

 \begin{figure*}[t!]
	\centering
		\includegraphics[width=1\textwidth]{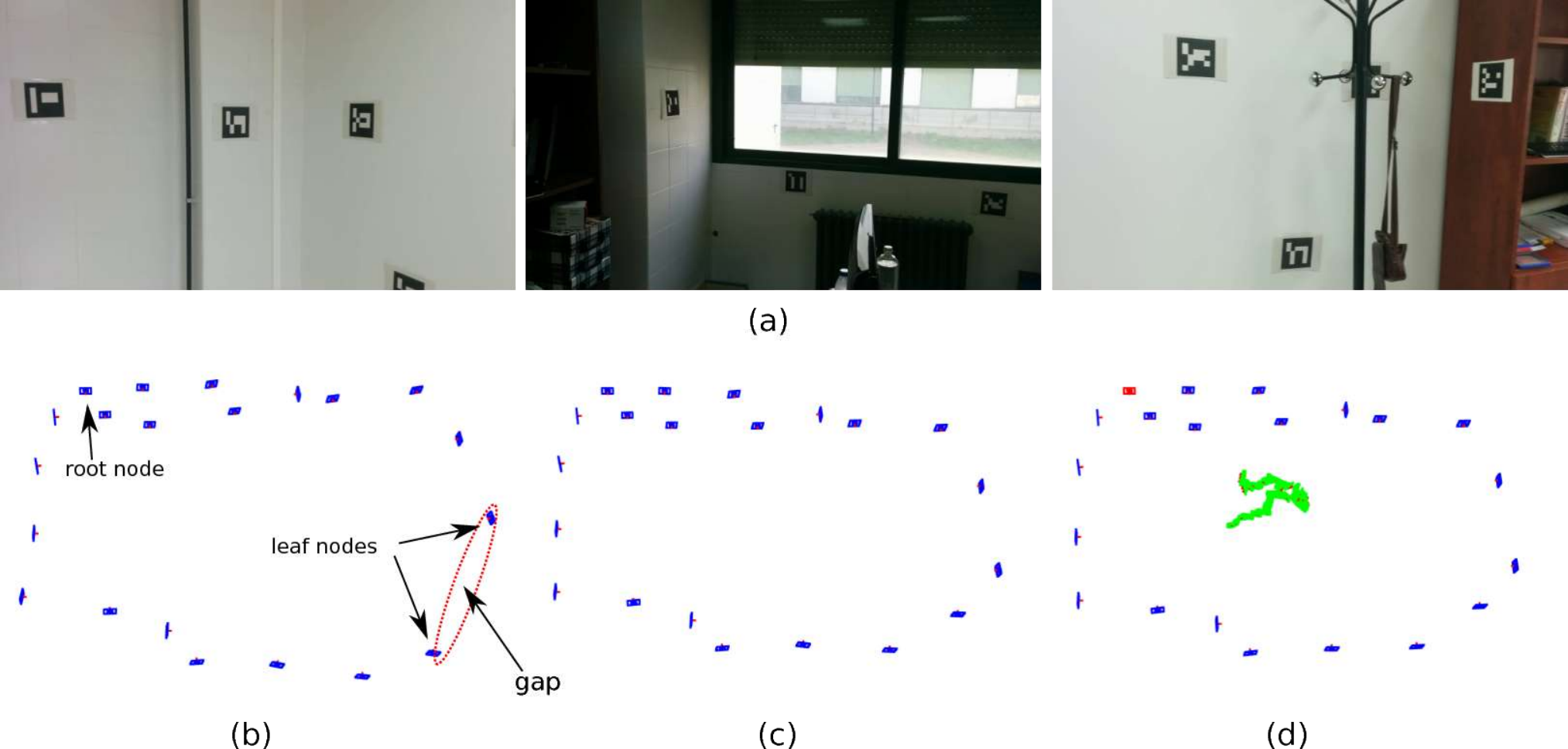}
		\caption{
	Results of the third test. (a) Snapshots of the recorded sequence. (b) Initial pose graph $G$. See the gap appearing between the last nodes of the spanning tree. (c) Optimized pose graph $\widetilde{G}$ obtained by distributing rotational and translational errors along the basis cycles. (d) Final marker and frame poses optimized. }
	\label{fig::office1}
\end{figure*}

\subsection{Second test: small box}

In the second test, we fixed  $32$ markers of  $1.2$~cm to a  cardboard box of dimensions  $16 \times 11 \times 9$~cm (see Fig. \ref{fig::arcube_test}(a)). This case is particularly difficult for the planar pose estimators because of  the small size of the markers  causes the ambiguity problem very often. The sequence was recorded with the Nexus-5 phone camera and has a total $640$ frames. Our method required $12$~secs to complete the processing  obtaining the results shown in Fig~\ref{fig::arcube_test}(b). A visual inspection of the results shows that the 3D location of all markers in the scene are correctly determined. However, in this test  it is not possible to obtain quantitative results since the ground truth is not available. 

The VisualSFM was also employed to process the matches obtaining the results shown in Fig~\ref{fig::arcube_test}(c). In this case the process  required
$68$~secs to complete and was only capable of finding $104$ out of the of the $128$ marker corners. In addition, a visual inspection of the trajectory shows higher errors than the results of our method. 

Finally, the results of OpenMVG  are shown Fig~\ref{fig::arcube_test}(d). In this case the processing required
$2508$~secs to complete, but it was capable of finding all marker corners and, as it can be observed, the camera trajectory  was successfully recovered.

\subsection{Third test: small room}
In the third test we placed $21$ markers of size $12.5$~cm wide onto the walls of a small office of approximately $6\times4$~meters. 
We recorded a sequence of $293$ frames and processed it with our method. Figure \ref{fig::office1}(a) shows some of the images recorded. The reconstruction took $2.25$~secs in our computer.  Figure \ref{fig::office1}(b) shows the marker poses estimated by the initial pose graph, i.e., without distributing the errors along the cycles of the graph (Sect.~\ref{sec:pose_graph_optimization}). In this example, the pose quiver is not fully connected and the maximum depth  of the {\it mst} is five. So, the small errors in  the path from the starting node to the leaf nodes of the spanning tree propagates creating the final gap shown in the Figure. The  reconstructed markers obtained after applying the graph optimization method are shown in Figure~\ref{fig::office1}(c). It is clear that the accumulative error  has been reduced. From this initial locations, the final optimized results are shown in Figure~\ref{fig::office1}(d), where the frame poses have been set out in green. In this case, we do not have ground truth, but a qualitative inspection reveals that the solution obtained resembles the true locations.

We also run the VisualSFM and OpenMVG tools on the matches of this sequence, however, they were incapable of reconstructing   the whole scene. This is because these   methods   need a higher number of matches between the frames in order to compute their relative poses (using either the fundamental matrix or the homographies). 

\subsection{Fourth test: laboratory reconstruction}

For this test we have placed a total of $90$ markers along our laboratory, which is comprised by two rooms  connected by a door. Each room has an approximated dimension of $7\times7$ meters. The laboratory was scanned using a Leica 3D laser scanner (see Fig~\ref{fig::lab12_3d}(a)) that provided a  3D point cloud from which we could manually select the   ground truth marker corners.

In this test, we recorded a sequence of $6998$ frames moving along the two rooms of the laboratory. The total time required to process the sequence was $185$~secs and it must be noticed that the final  number of variables to optimize in Eq.~\ref{eq::total_repr_err} was  $42,555$. The results of the different steps of  our algorithm can be seen in Fig~\ref{fig::lab12_3d}(b-e).
Figure~\ref{fig::lab12_3d}(b) shows the initial pose graph, where we have enclosed  in red ellipses the regions corresponding to the leaf nodes of the spanning tree. It is clear, specially in room 1, that the initial pose graph is not capable of creating a consistent reconstruction. In Figure~\ref{fig::lab12_3d}(c) we have shown the final results of optimizing Eq.~\ref{eq::total_repr_err} using as initial estimation the one shown in  Figure~\ref{fig::lab12_3d}(b), i.e., the  results of optimized Eq.~\ref{eq::total_repr_err} without applying the graph optimization method of Sect.~\ref{sec:pose_graph_optimization}. As can be seen, the LM algorithm is not capable of finding a good result from that initial solution. 
Figure~\ref{fig::lab12_3d}(d) shows the  reconstruction obtained after applying the pose graph optimization  proposed. And finally, Figure~\ref{fig::lab12_3d}(e) shows the final reconstruction along with the camera trajectory coloured in green. The ACE obtained in the localization of the markers for this test was $2.1$~cm. The ATE cannot be computed since there is no ground truth for this sequence.

When the data was processed using the SfM tools, they was incapable of reconstructing the scene.

\subsection{Fifth test: laboratory reconstruction under rotational movement }

\begin{figure*}[t!]
	\centering
		\includegraphics[width=0.9\textwidth]{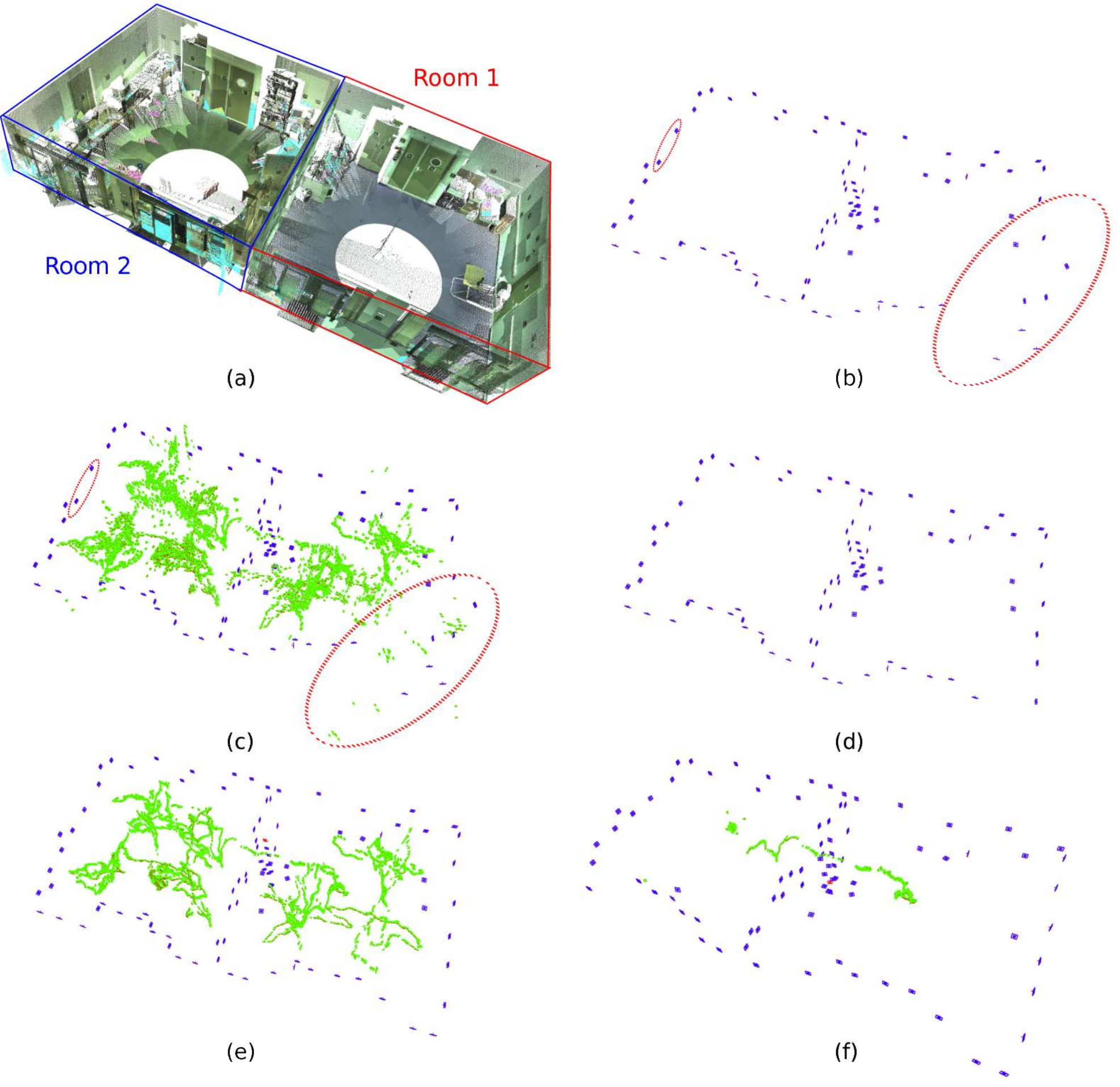}
	\caption{ Results of  fourth and fifth tests.   (a) Three-dimensional reconstruction of the laboratory using a Leica laser scanner. (b) Initial pose graph for the fourth test. Notice the erroneous reconstruction for the leaf nodes of the minimum spanning tree enclosed in red ellipses. (c) Optimization result when using the solution in {\it (b)} as starting solution. Notice that errors can not be solved and the LM algorithm finishes in a local minimum. (d) Pose  graph after error distribution along the cycles. (e) Final optimization result of the LM algorithm when starting from {\it (d)}. (f) Results of the fifth test, in which the camera undergo mostly rotational movements.}
	\label{fig::lab12_3d}
\end{figure*}

This test is aimed at testing the reconstruction capabilities of our method when the camera undergo mostly rotational movements. We placed the camera in the center of the first room and rotated $360$~deg. Then, the camera was moved to the center of the second room, where again, it was rotated $360$~deg. The sequence has $2103$ frames and the reconstruction required $66$~secs of computing time. In this case,  only $76$ out of the $90$ markers were visible from the recorded locations, and  the ACE obtained was $2.9$~cm.  The final reconstruction is shown in Figure~\ref{fig::lab12_3d}(e). Again,  the  SfM tools  were incapable of reconstructing the scene.

\subsection{Sixth test: comparison with SLAM systems }

So far, we have reported the comparison of our method with SfM  approaches. In this section, we compare against  two  Simultaneous Localization And Mapping approaches: LSD-SLAM~\cite{engel14eccv} and ORB-SLAM\footnote{We employed the latest version of the software ORB-SLAM2.}~\cite{orb-slam}.  The first method is based on dense stereo matching, while the second one relies on ORB keypoints, however,  both are capable of managing loop closures. 

We recorded three video sequences in room 1 using a PtGrey FLEA3 camera, which recorded frames of $2080\times1552$ pixels at a frame rate of $30$~Hz. The camera ground truth locations were recorded using an Optitrack motion capture system, that tracks camera poses at $120$~Hz using six cameras placed around room 1.      For the LSD-SLAM method, we resized images to $640\times480$ as suggested by their authors.

\begin{figure}[t!]
	\centering
		\includegraphics[width=0.5\textwidth]{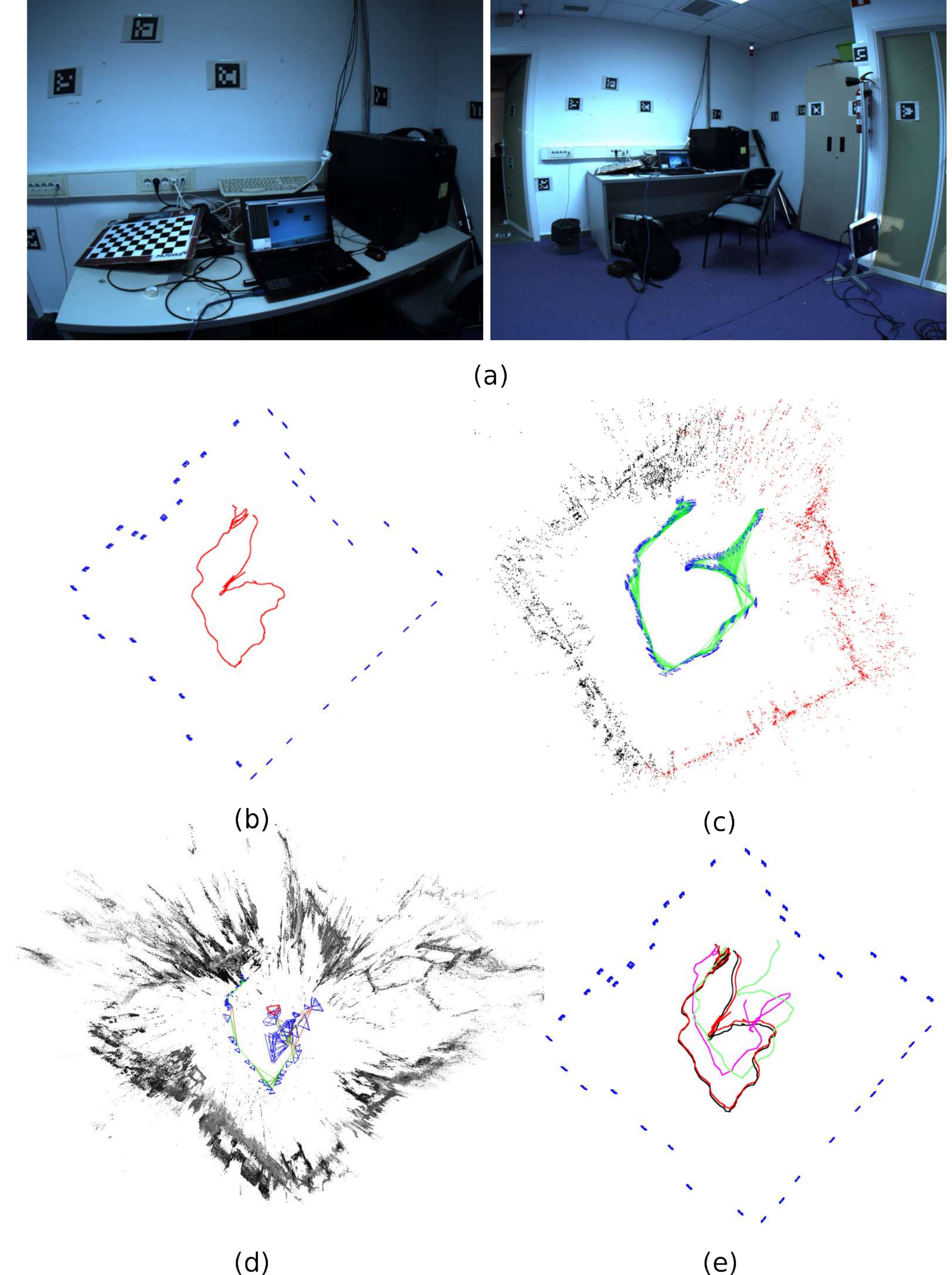}
	\caption{Results for the  {\it SLAM-Seq 1} sequence of the sixth test. (a) Initial and final images of {\it SLAM-Seq 1}. The same part of the room  is visible in both images. (b) Marker reconstruction  and trajectory obtained with our method. (c) Three-dimensional reconstruction of ORB-SLAM2 along with the trajectory estimated. (d)  Three-dimensional reconstruction of LSD-SLAM along with the trajectory estimated. (e)  Comparison of the estimated trajectories. Black line: ground truth. Red line: our method. Green line: LSD-SLAM. Pink Line: ORB-SLAM2.}
	\label{fig::slam_1}
\end{figure}

In the three sequences, the camera was moved around the room pointing to the walls. The video started and finished pointing at the same region of the room, so that the closure of the loop  could be detected. Although pointing at the same spot, the initial and final locations were separated by two meters approximately. Figure~\ref{fig::slam_1}(a) shows the initial and final images of one of the sequences.

For evaluation purposes, we employed the ATE by comparing the trajectory provided by each method with the ground truth provided by the Optitrack system.  Since both SLAM methods calculate the trajectories in an unknown scale, it was first necessary to find the best scale in order to compare the results. To that end, we did a grid search into the range $[0.01,3]$ at steps of $0.001$.

The results of the different methods are shown in the Table~\ref{tab::slam_ate}, where the ATE of each method is reported. As can be observed, our method outperforms in all the tested sequences.  To graphically show the results, we draw attention to Figure~\ref{fig::slam_1}(b-e)  that shows the results obtained for the test sequence {\it SLAM-Seq 1}.  Figure~\ref{fig::slam_1}(b) shows in blue  the three-dimensional marker reconstruction of our method, and as a red  coloured line,  the estimated frame poses. Then ACE in the reconstruction of the markers corners was $1.5$~cm for that sequence.

Figure~\ref{fig::slam_1}(c) shows the reconstruction results obtained by the ORB-SLAM2 method, where the selected keyframes are printed in blue. Finally, Figure~\ref{fig::slam_1}(d) shows the reconstruction obtained by the LSD-SLAM method, and again the keyframes are shown in blue. Finally, Figure~\ref{fig::slam_1}(e) shows the trajectories of the three methods along with the ground truth trajectory obtained by the Optitrack system. The black line corresponds to the ground truth, the red one is the result of our method, while green and pink correspond to the LSD-SLAM and ORB-SLAM2 respectively. As it can be observed, our method is capable  of calculating in this sequence a much better approximation of the frame poses than the other two methods.

\begin{table}
\begin{center}
 \caption{Absolute trajectory error (in meters) for the three sequences registered with a motion capture system. The table shows the results of our method and the monocular SLAM methods employed for comparison. As can be observed, our method obtains the best results in the three sequences. }
\begin{tabular}{ l|c|c|c}
  \textit{Sequence}   & Ours & LSD-SLAM & ORB-SLAM2 \\
  \hline
  SLAM-Seq 1 & \textbf{0.0447}~m & 0.440~m &0.231~m \\
  SLAM-Seq 2 & \textbf{0.0433}~m & 0.117~m &0.913~m \\
  SLAM-Seq 3 & \textbf{0.0694}~m & 0.652~m& 0.314~m \\

\end{tabular}

  \label{tab::slam_ate}
  \end{center}
\end{table}

\subsection{Seventh test: minimal configuration }
\begin{figure*}[t!]
	\centering
		\includegraphics[width=1\textwidth]{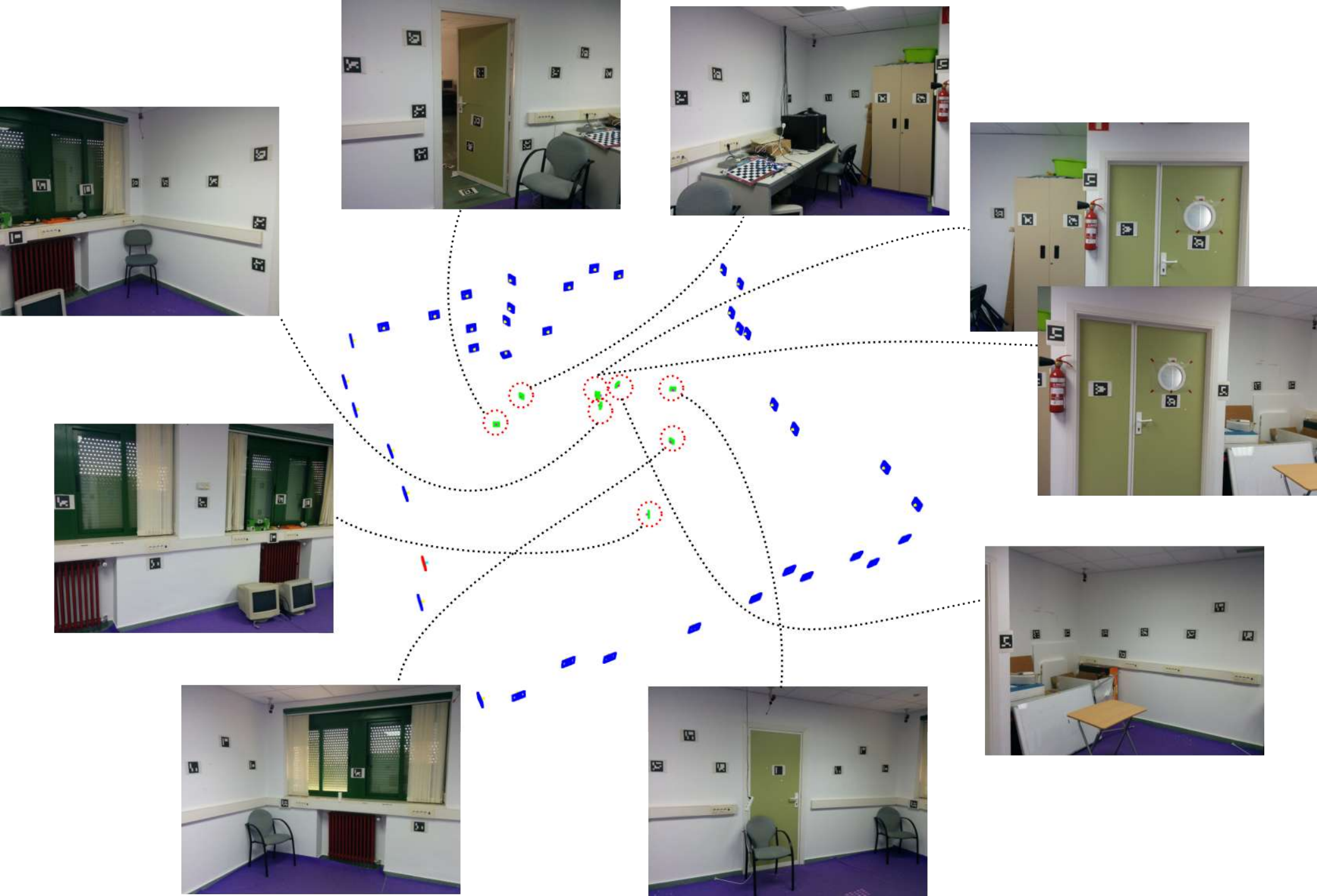}
	\caption{Results obtained for the Seventh test. Marker poses of Room 1 are reconstructed using only $9$ images of the environment.}
	\label{fig::minimal_test}
\end{figure*}

This final test aims at evaluating the reconstruction capabilities of the proposed method using a minimal set of images of the environment. For that purpose, we have taken seven pictures of the room 1 using the camera of Nexus 5 mobile phone at a resolution of $3268\times2448$ pixels. The pictures along with the reconstruction obtained are shown in Fig.~\ref{fig::minimal_test}. The computing time required was $1.7$~secs and the ACE obtained for the reconstructed markers is $2.21$~cm. For this test, we do not have the ground truth camera locations so the ATE cannot be computed. 

In our opinion, the results of this test show that  squared planar markers are a very convenient approach  for camera localization in controlled environments requiring  very limited number of views to obtain a very precise map of the environment.
In contrast,  keypoint-based SLAM approaches would require a much higher number of views in order create a reliable map that could be used to reliably localize a camera.

\section{Conclusions}
\label{sec::conclusions}
This paper has proposed a novel approach for mapping and localization using squared planar markers. The method runs offline on a video sequence by first collecting all available observations to create a  quiver with the relative  poses of the  observed markers. Then, an initial pose graph is created that is later refined by distributing the rotation and translational errors around the cycles. Using the initial marker poses from the refined graph, an initial estimation of the frame poses are obtained considering the possibility of ambiguity. Finally, all poses are refined using a Levenberg-Marquardt optimization to reduce the reprojection error of the marker corners in all observed frames. The proposed optimization function  ensures that the markers geometry is kept during the optimization process.

The proposed method has been compared with Structure from Motion and Simultaneous Localization and Mapping techniques based on keypoints. The results show that our method is capable of obtaining better maps and localization results under a wider range of viewpoints.

\section{Acknowledgments}
This project has been funded under projects TIN2016-75279-P and  RTC-2016-5661-1.

\bibliographystyle{elsarticle-num}

\end{document}